\documentclass[letterpaper, 10 pt, journal, twoside]{IEEEtran} 

\IEEEoverridecommandlockouts                              

\usepackage{graphics} 
\usepackage{graphicx}
\usepackage{amsmath} 

\usepackage{amssymb}  
\usepackage{booktabs}
\usepackage{algorithm}
\usepackage{algorithmic}
\usepackage{url}
\usepackage{color,soul}
\usepackage{tabularx}

\usepackage[font=footnotesize,labelfont=bf]{caption}
\begin{document}

\title{\LARGE \bf
AirCapRL: Autonomous Aerial Human Motion Capture \\ using Deep Reinforcement Learning
}


\author{Rahul Tallamraju$^{1}$, Nitin Saini$^1$, Elia Bonetto$^1$, Michael Pabst$^1$, Yu Tang Liu$^1$, \\ Michael J.\ Black$^1$ and Aamir Ahmad$^{1,2}$%
\thanks{Manuscript received: February, 24, 2020; Revised June, 1, 2020; Accepted July, 12, 2020.}
\thanks{This paper was recommended for publication by Editor Tamim Asfour upon evaluation of the Associate Editor and Reviewers' comments.} 
\thanks{$^{1}$Max Planck Institute for Intelligent Systems, T\"ubingen, Germany.
        {\tt\footnotesize {firstname.lastname}@tuebingen.mpg.de}}%
\thanks{$^{2}$Department of Aerospace Engineering and Geodesy, University of Stuttgart, Germany.}
\thanks{Digital Object Identifier (DOI): see top of this page.}
}

\markboth{IEEE Robotics and Automation Letters. Preprint Version. Accepted July, 2020}
{Tallamraju \MakeLowercase{\textit{et al.}}: AirCapRL: Autonomous Aerial Human Motion Capture using Deep Reinforcement Learning}

\maketitle

\begin{abstract}  
In this letter, we introduce a deep reinforcement learning (DRL) based multi-robot formation controller for the task of autonomous aerial human motion capture (MoCap). We focus on vision-based MoCap, where the objective is to estimate the trajectory of body pose and shape of a single moving person using multiple micro aerial vehicles.
State-of-the-art solutions to this problem are based on classical control methods, which depend on hand-crafted system and observation models. Such models are difficult to derive and generalize across different systems. Moreover, the non-linearities and non-convexities of these models lead to sub-optimal controls.
In our work, we formulate this problem as a sequential decision making task to achieve the vision-based motion capture objectives, and solve it using a deep neural network-based RL method.
We leverage proximal policy optimization (PPO) to train a stochastic decentralized control policy for formation control.
The neural network is trained in a parallelized setup in synthetic environments.
We performed extensive simulation experiments to validate our approach. Finally, real-robot experiments demonstrate that our policies generalize to real world conditions. 
\end{abstract}


\begin{IEEEkeywords}
Reinforecment Learning; Aerial Systems: Perception and Autonomy; Multi-Robot Systems; Visual Tracking.
\end{IEEEkeywords}

\IEEEpeerreviewmaketitle


\section{Introduction}

\IEEEPARstart{H}{uman} motion capture (MoCap) implies accurately estimating 3D pose and shape trajectory of a person. 3D pose, in our case, consists of the 3D positions of the major human body joints. Shape is usually parameterized by a large number (in thousands) of 3D vertices. In a laboratory setting MoCap is performed using a large number of precisely calibrated and high-resolution static cameras. To perform human MoCap in an outdoor setting or in an unstructured indoor environment, the use of multiple and autonomous micro aerial vehicles (MAVs) has recently gained attention \cite{MarkerlessNitin19,ActiveTallamraju19, DeepPrice18,nageli2018flycon,xu2018flycap}. Aerial MoCap of humans/animals facilitates several important applications, e.g., search and rescue using aerial vehicles, behavior estimation for endangered animal species, aerial cinematography and sports analysis.

Realizing an aerial MoCap system involves several challenges. The system's robotic front-end \cite{ActiveTallamraju19} must ensure that the subject i) is accurately and continuously followed by all aerial robots, and ii) is within the field of view (FOV) of the cameras of all robots. The back-end of the system estimates the 3D pose and shape of the subject, using the images and other data acquired by the front-end \cite{MarkerlessNitin19}. The front-end poses a formation control problem for multiple MAVs. In this letter, we propose a deep neural network-based reinforcement learning (DRL) method for this formation control problem.

\begin{figure}[!t]
 \includegraphics[width=\columnwidth]{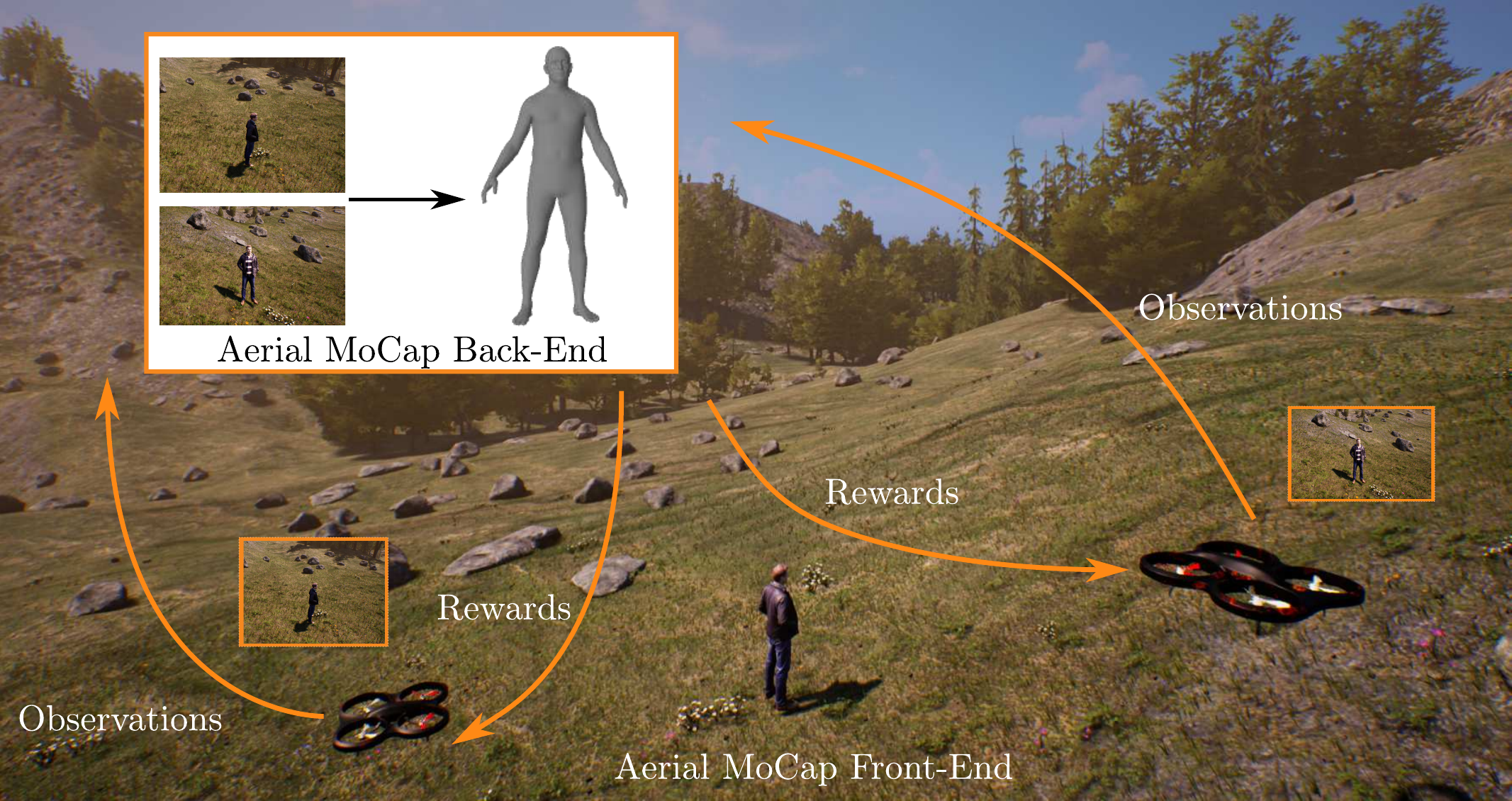}
 \caption{An illustration of an aerial MoCap system where MAV agents learn formation control policies based on MoCap performance rewards.}
 \label{fig:cover}
\end{figure}

Below, we describe the drawbacks in state-of-the-art methods and highlight the novelties in our work to address them.

In existing solutions \cite{MarkerlessNitin19,ActiveTallamraju19, DeepPrice18} the front and back end are developed independently -- the formation control algorithms of the existing aerial MoCap front ends assume that the person should be centered in every MAV's camera image and she/he should be within a threshold distance to each MAV. These assumptions are intuitive and important. Also, experimentally it has been shown that it leads to a good MoCap estimate. However, it remains sub-optimal without any feedback from the estimation back-end of the MoCap system. The estimated 3D pose and shape are strongly dependent on the viewpoints of the MAVs. In the current work, we take a learning-based approach to map and embed this dependency within the formation control algorithm. This is our \textbf{first key novelty}.   

Existing approaches \cite{ActiveTallamraju19, DeepPrice18,nageli2018flycon,xu2018flycap} depend on tediously obtained system and observation models. State-of-the-art solutions to formation control problems, which involve perception-related objectives, derive observation models for the robot's camera and the desired subject to compute real-time robot trajectories \cite{ActiveTallamraju19, PampcFALANGA18, RealNageli2017}. Since these observation models are based on assumptions on the shape and motion of the subject, sensor noise and the system kinematics, the computed trajectories are sub-optimal. We overcome the aforementioned issue by addressing the formation control for aerial MoCap as a multi-agent reinforcement learning (RL) problem. This is the \textbf{second key novelty} of our approach. We let the MAVs learn the best control action given only the subject perception observable through the MAV's on-board camera images, without making any assumptions on the observation model. 

The key insights which enable us to do this are i) the sequential decision making nature of the formation control problem with MoCap objectives, and ii) the feasibility of simulating control policies in synthetic environments. We leverage the actor-critic methodology of training an RL agent with a centralized training and decentralized execution paradigm. At test time, each agent runs a decentralized instance of the trained network in real-time. 
We showcase the performance of our method in several simulation experiments. We evaluate the quality of the generated robot trajectories using the pose and shape estimation algorithms in \cite{kolotouros2019spin}, \cite{liang2019shape} and \cite{MarkerlessNitin19}. Additionally, we compare our new approach with the state-of-the-art model-based controller from \cite{ActiveTallamraju19}. A demonstration and comparison with the method of \cite{ActiveTallamraju19} on a real MAV is also presented.
Code and implementation details of our method is provided in the supplementaty material. 


\section{Related Work}

\textit{Aerial Motion Capture Methodologies:}
A marker-based multi-robot aerial motion capture system is presented in \cite{nageli2018flycon}. Here, pose of the person and the robots are jointly estimated and optimized online. A multi-robot model-predictive controller is used to compute trajectories which optimize the camera viewing angle and person visibility in the image. Marker-based methods suffer from tedious setup times, and optimal control methods for trajectory following can lead to sub-optimal policies for motion capture due to perceptual objectives.
A markerless aerial motion capture system using multiple aerial robots and depth cameras is proposed in \cite{xu2018flycap}. They use a non-rigid registration method to track and fuse the depth information from multiple flying cameras to jointly estimate the motion of a person and the cameras. Their approach works only indoors and the initial registration step can take a long time similar to other marker based method setups. In one of our previous works, \cite{saini2019markerless}, we introduced a vision-based (monocular RGB) markerless motion capture method using multiple aerial robots in outdoor scenarios. The pose and shape of the subject and the pose of the cameras are jointly estimated and optimized in \cite{saini2019markerless}. While our other previous work \cite{ActiveTallamraju19} introduces a front-end of our outdoor aerial MoCap system, \cite{saini2019markerless} describes the back-end. 

\textit{Perception-Aware Optimal Control Methods for Target Tracking:}
In \cite{PampcFALANGA18}, a perception-aware MPC generates real-time motion plans which maximize the visibility of a desired static target. 
In \cite{lee2020aggressive} a  deep learned optical flow algorithm and non-linear MPC are jointly utilized to optimize a general task-specific objective. The optical flow dynamics are explicitly embedded into the MPC to generate policies which ensure the visibility of target features during navigation. 
An occlusion-aware moving target following controller is proposed in \cite{jeon2019online}. Here, metrics for target visibility are utilized to navigate towards a moving target and constrained optimization is leveraged to navigate safely through corridors.
In the above works, the motion plans are generated only for a single aerial robot to track a single generic target.  
In our previous work \cite{ActiveTallamraju19}, a non-linear MPC based formation controller for active target perception is introduced for target following. The controller assumes Gaussian observation models and linearizes system dynamics. Using these, it identifies a collision-free trajectory which minimizes the fused uncertainty in target position estimates.
In contrast to that, in our current work we learn a control policy to explicitly improve the quality of 3D reconstruction of human pose. An implicit perception-aware target following behavior evolves out of the controller for both single and multi-agent scenarios.

\textit{Learning based Control for Aerial Robots for Perception Driven Tasks:}
Optimal control methods are computationally expensive, require explicit estimation of the state of the system and world, and depend mostly on hand-crafted system and observation models. Thus, it can often lead to sub-optimal behaviors.
A model-predictive control guided policy search was proposed in \cite{zhang2016learning} where supervised learning is used to obtain policies which map the on-board aerial robot sensor observations to control actions. The method does not require explicit state estimation at test time and plans based on just input observations.
In \cite{bonatti2019autonomous} authors used a deep Q-learning based approach for cinematographic planning of an aerial robot (or MAV). A discrete action policy was trained on rewards that exploit aesthetic features in synthetic environments. User studies were performed to obtain the aesthetic criteria. In contrast to that, our current work proposes single and multi-agent MAV control policies that reward the  minimization of errors in body pose and shape estimation.
A proximal policy optimization (PPO) based distributed collision avoidance policy was proposed in \cite{fan2018fully}. A centralized training and decentralized execution paradigm was leveraged to obtain a policy that maps laser range scans to non-holonomic control actions. 
In \cite{Everett2018} the authors propose an A3C actor-critic algorithm to develop reactive control actions in dynamic environments. Each agent's ego observations and an LSTM-encoded dynamic environmental observations are inputs to a fully connected network. Their goal is to obtain a fully distributed control policy.
In contrast to the aforementioned works, we propose a model-free deep reinforcement learning approach to the MoCap-aware MAV formation control problem. In our work, a policy neural network directly maps observations of the target subject to control actions of each MAV without any underlying assumptions of the observation model or system dynamics. 

\section{Methodology}

\subsection{Problem Statement}
\begin{figure}
	\centering
	\includegraphics[scale=0.60]{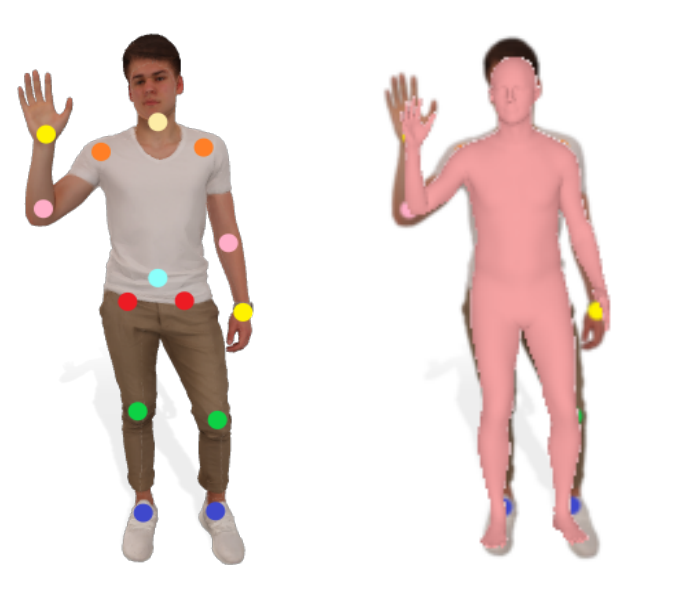}
\caption{The ground truth body joints (left) and estimated pose and shape overlaid (right).}
\label{fig:mesh}
\end{figure}

Let there be a team of K MAVs (with quadcopter-type dynamics) tracking a person P. The pose of the $k^{th}$ MAV in the  world frame at time $t$ is given by $\xi_t^{k} = [(\mathbf{x}_t^{k})^\top ~ (\Theta_t^{k})^\top] \in \mathbb{R}^6$, where $(\mathbf{x}_t^{k})^\top$ denotes the 3D position of the MAV's center in Cartesian coordinates and $(\Theta_t^{k})^\top$ denotes its orientation in Euler angles. Each MAV has an on-board, monocular, perspective camera. It is important to note that the camera is rigidly attached to the MAV's body frame, pitched down at an angle of $\theta_\textrm{cam}$. The global pose of the person is given by $\xi_t^P = [(\mathbf{x}_t^P)^\top ~ (\Theta_t^P)^\top ~(\mathbf{x}_{j,t}^P~\forall~j; j=1 \cdots 14)^\top ] \in \mathbb{R}^{48}$. $(\mathbf{x}_t^P)^\top$ and $(\Theta_t^P)^\top$ are the body's 3D center and global orientations, respectively. $\mathbf{x}_{j,t}^P$ denotes the 3-D position of a joint $j$ from a total of fourteen joints considered for the MoCap of the subject. Ground truth joints considered are visualized as circles in Fig.~\ref{fig:mesh}. The MAVs operate in an environment with neighboring MAVs as dynamic obstacles. Their task is to autonomously fly and record images of the person using their on-board camera. The formation control goal of the MAV team is to cooperatively navigate in a way such that the error in 3D pose estimates of the subject is minimized. 

\subsection{Formulation as a Sequential Decision Making Problem}

Intuitively, the accuracy of aerial MoCap depends on the following two factors.
\begin{itemize}
 \item The subject should always remain completely in the FOV of every MAV's camera, occupying maximum possible area on the image plane.
 \item The subject is visually encapsulated from all possible directions (viewpoints).
\end{itemize}

Based on these intuitions and experimentally derived models for single and multiple camera-based observations, in our previous work \cite{ActiveTallamraju19} we approached this problem using a model predictive control (MPC) based formation controller. The MPC objective was to keep a threshold distance to the subject while satisfying constraints that enable uniform distribution of viewpoints around the subject. Additionally, a yaw controller ensured that the subject was always centered on the image plane. As discussed in the introduction, this method is hard to generalize because to i) it is agnostic to how the 3D pose and shape was estimated by the back end, and ii) it needs carefully derived observation models.

To address these issues in this work we take a deep reinforcement learning-based approach. We model this formation control problem as a sequential decision making problem for every MAV agent. Dropping the MAV superscript $k$, for each agent the problem is defined by the tuple $(S,O,A,T,R)$, where $S$ is the state-space, $O$ is the observation-space, $A$ is the action-space, $T$ is the environment transition model, and $R$ is the reward function. At each time instance $t$, an agent at state $s_t$ has access to an observation $o_t$ using its cameras and on-board sensors. The agent then chooses an action $a_t$, which is conditioned on $o_t$ using a stochastic policy $\pi_\theta(a_t|o_t)$, where $\theta$ represents parameters of a neural network. The agent experiences an instantaneous reward $r_t(s_t,a_t)$ from the environment indicating the goodness of the chosen action. We approach the problem without any underlying assumptions or knowledge about the environment transition model $T$. To this end, we leverage a model-free deep reinforcement learning method to train the agents. We will further describe the states, observations and actions in detail. Due to ease of notations and to keep the RL training computationally tractable, we will consider 2 MAV agents in this letter, i.e, $K=2$.  Rewards are described later when we discuss our proposed methodology in sub-section~\ref{subsec:proposed_method}.

\subsubsection{States and Observations}
Each agent's environment state, $s_t$, includes the MAV pose $\xi_t$, its neighboring MAV's pose $\bar{\xi}_t$ and the MoCap subject's pose $\xi_t^P$.
\begin{equation}
 s_t = [\xi_t ~~ \bar{\xi}_t ~~ \xi_t^P];  ~~ o_t = [\mathbf{y}_t^{P} ~~ \mathbf{\dot{y}}_t^{P} ~~ \psi_t^{P} ~~ \mathbf{y}_t^{N} ~~ \psi_t^{P,N}]
 \label{eqn:state_obs}
\end{equation}

The observation vector $o_t$ is given by (\ref{eqn:state_obs}). Its first two components are the measurements of the person $P$'s position and velocity made by the agent in its local Cartesian coordinates. This is given by $[\mathbf{y}_t^{P} ~~ \mathbf{\dot{y}}_t^{P}] \in\mathbb{R}^6$. The third component of the observation vector is the measurement of the relative yaw orientation of the person with respect to the robot's global yaw orientation, denoted by $\psi_t^{P}$.  Here we emphasize that we make no assumptions regarding the uncertainty model associated with these measurements. However, we assume that these measurements are available using a vision-based detector. In our synthetic training environment we directly use the available ground truth position and orientation of the person and the MAV to compute these measurement. In real robot scenarios we use Vicon readings to calculate it. The fourth component is the 3D position measurements to the neighboring MAV agent in the local Cartesian coordinates of the observing agent. This is given by $\mathbf{y}_t^{N} \in\mathbb{R}^3$. The fifth component is the measurement of the relative yaw angle orientation of the person with respect to the neighboring robot's global yaw orientation, denoted by $\psi_t^{P,N}$.


\subsubsection{Actions}
\label{subsec:actions}

Action $a_t$ is sampled from the control policy $\pi_\theta(a_t|o_t)$ for an input observation $o_t$. In our formulation, actions consist of egocentric 3-D linear translational velocity of the agent, given by $\mathbf{v}_t = [{vx}_t ~~ vy_t ~~ vz_t]$ and a rotational velocity $\omega_t$ about its z-axis. The chosen action defines a way-point $\{\mathbf{x}^w_t,~\phi^w_t\}$, which is obtained as $~\mathbf{x}^w_t = \mathbf{x}_t + \mathbf{R}(\phi_t)\mathbf{v}_t\Delta, ~\phi^w_t=\phi_t+\omega_t\Delta$, for the agent in the world frame. $\{\mathbf{x}^w_t,~\phi^w_t\}$ is provided to low-level geometric tracking controller (Lee controller) \cite{lee2010geometric} of the agent. $\mathbf{x}_t$, as defined before, denotes the current 3D position of the agent. $\mathbf{R}(\phi_t)\in SO(3)$ is a rotation matrix. Thus,
\begin{equation}
 a_t= [\mathbf{v}_t ~~ \omega_t] \in\mathbb{R}^4
 \label{eqn:actions}
\end{equation}


\subsection{Proposed Methodology}
\label{subsec:proposed_method}


Training multiple agents to achieve multiple objectives is a complex and computationally demanding task. In order to have a systematic comparison we first develop our approach for a single agent case and then for multi-agent scenario. Meaning, we train (and then evaluate and compare) two different kinds of agents, and hence, networks. These are i) a single agent with only MoCap objectives, and ii) multi-agents (2 in our case) with both MoCap and collision avoidance objectives.

%

We hypothesize that, using the first kind of network, an agent will learn to follow the person and orient itself in the direction of the person in order to achieve accurate MoCap from the back-end estimator. On the other hand, using the second network, the agents will learn how to avoid each other and distribute themselves around the person to cover all possible viewpoints. We also hypothesize that the best navigation policies for the robot(s) for the MoCap task should significantly depend on the MoCap's accuracy-related rewards, while other rewards may or may not be required.


\subsubsection{Network 1: Single Agent Network}
All variants of single agent network use the following states and observations, where the superscript 1 denotes single agent network. 
\begin{equation}
 s_t^1 = [\xi_t ~~ \xi_t^P]; ~~ o_t^1 = [\mathbf{y}_t^{P} ~~ \mathbf{\dot{y}}_t^{P} ~~ \psi_t^{P}]
\end{equation}
The actions for all single agent network variants consist of $a_t$ as stated in (\ref{eqn:actions}). They are all trained on a moving subject. These variants differ only in their reward structure as described below. The rewards are computed at every timestep. However, for sake of clarity we drop the subscript $t$ from the reward variables.

\paragraph{Network 1.1 -- Only Centering Reward} 
In this variant we only reward the agent based on the intuitive reasoning of keeping the person as close as possible to the center of the image from the MAV agent's on-board camera. It is calculated as follows. 
\begin{equation}
 r_{\textrm{center}} = 1 - \tanh({c_1d_{\mathrm{px}}}),
 \label{eqn:centrtingreward}
\end{equation}where $d_{\mathrm{px}}$ is the distance between the center of the person's bounding box on the image to the image center, measured in pixels. $c_1=0.01$ is a weighting constant. Note that keeping the person centered in each frame is not the goal of this work. As per the above-stated hypothesis, centering reward may not be required at all. Thus, Network 1.1 will only serve as a comparison benchmark to highlight that a MoCap's accuracy-related reward is explicitly required.

\paragraph{Network 1.2 -- SPIN Reward}
In this variant of the network we reward the agent based on the output accuracy of the MoCap back end. For this, we use SPIN \cite{kolotouros2019spin}, a state-of-the-art method for human pose and shape estimation using monocular images. At every time-step of training, we use SPIN on the image acquired by the agent and compute an estimate of $\mathbf{\hat{x}}_{j,t}^P \forall j ; j=1\cdots14$ corresponding to all 14 joints. In the synthetic training environment we have access to the true values of these joints, denoted by, $\mathbf{\bar{x}}_{j,t}^P \forall j ; j=1\cdots14$. SPIN reward is then given by
\begin{equation}
 r_{\textrm{SPIN}} = 1 - \tanh({c_2d_{\mathrm{J}}}),
\label{eqn:spinreward}
\end{equation}where $d_{\mathrm{J}} = \frac{1}{14}\sum_{j=1}^{14}(||\mathbf{\hat{x}}_{j,t}^P  - \mathbf{\bar{x}}_{j,t}^P||_2)$ and $c_2 = 5$ is a weighting constant.




\paragraph{Network 1.3 -- Weighted SPIN Reward}
Network 1.2 rewards the agent equally for the accuracy of each joint. However, the joints further away from the pelvis (also mentioned as the root joint), like hands or foot, have a greater tendency to be in an erratic motion than the ones closer to the root, like hips. To account for this, in the network variant 1.3 we penalized the outward joints more and hence define a Weighted SPIN reward as,
\begin{equation}
 r_{\textrm{WSPIN}} = 1 - \tanh({c_2d_{\mathrm{W}}}), 
 \label{eqn:weightedspinreward}
\end{equation}where
$d_{\mathrm{W}} = \frac{1}{14}\sum_{j=1}^{14}(w_j||\mathbf{\hat{x}}_{j,t}^P  - \mathbf{\bar{x}}_{j,t}^P||_2)$
and $w_j$s are positive weights that sum to 1.

\paragraph{Network 1.4 -- Centering and Weighted SPIN Reward}
The last variant of the single agent uses a summed reward given as $r_{\textrm{sum}} = r_{\textrm{center}} + r_{\textrm{WSPIN}}$.

\smallskip

\subsubsection{Network 2: Multi-Agent Network}
All three variants of the multi agent network, described below, use the state as defined in (\ref{eqn:state_obs}).
The observations for Network variants 2.1 and 2.2 are equal to (\ref{eqn:state_obs}) without  $\mathbf{\dot{y}}_t^{P}$ as these variants are trained on a static subject. In these two variants the action space excludes yaw control. Hence, during their training, we use a separate yaw controller to always orient the agent towards the person. On the other hand, Network 2.3 is trained with the full observation space as stated in (\ref{eqn:state_obs}) on a moving subject, and it uses the full action space is as stated in (\ref{eqn:actions}). Meaning, Network 2.3 also includes yaw-rate control.

The difference in the reward structure is described below.

\paragraph{Network 2.1: Centering, collision avoidance and AlphaPose Triangulation Reward (Trained with Static Subject)} 
In this variant we use a sum of three rewards $r_{\textrm{center}}$, $r_{\textrm{col}}$ and $r_{\textrm{triag}}$. Here, $r_{\textrm{center}}$ is same as defined in (\ref{eqn:centrtingreward}). $r_{\textrm{col}}$ rewards avoiding collisions by penalizing based on the distance from the neighboring robot. It is computed as
\begin{equation}
r_{\textrm{col}} = \begin{cases}
-1 ,& \text{if } \|\mathbf{x}_t-\bar{\mathbf{x}}_t\|_2\geq \mathbf{x}_{thresh}\\
0.2 ,              & \text{otherwise}
\end{cases}
\label{eqn:colreward}
\end{equation}where $\mathbf{x}_{thresh} = 3$m in our implementation.

$r_{\textrm{triag}}$ is a simplified MoCap-specific reward in a 2-agent scenario, which we obtain using a triangulation-based method. AlphaPose \cite{cao2017realtime} is a state-of-the-art human joint detector which provides body joint detections on monocular images. At every time step we use it on the images obtained by the agent and its neighbor to obtain $o_{j,t}\in\mathbb{R}^{14}$ and $\bar{o}_{j,t}\in\mathbb{R}^{14}$, respectively. Using known camera intrinsics and extrinsics (from self-pose estimates) for both agents, a point in the image plane and its corresponding view from another camera, we can estimate the 3-D position of the point using a least squares formulation (equation (14.42) in \cite{prince2012computer}). Therefore, by using $o_j$ and $\bar{o}_{j,t}$, we estimate the 3D positions of all 14 joints of the subject as $\mathbf{\tilde{x}}_{j,t}^P \forall j; j=1\cdots14$ and compare it to ground-truth joint positions $\mathbf{\bar{x}}_{j,t}^P \forall j; j=1\cdots14$. Thus, $r_{\textrm{triag}}$ is given by
\begin{equation}
 r_{\textrm{triag}} = 1 - \tanh({c_3d_{\mathrm{triag}}}),
 \label{eqn:triagreward}
\end{equation}where $d_{\mathrm{triag}} = \frac{1}{14}\sum_{j=1}^{14}(||\mathbf{\tilde{x}}_{j,t}^P  - \mathbf{\bar{x}}_{j,t}^P||_2)$.


\begin{figure}[!t]
 \includegraphics[width=0.9\columnwidth]{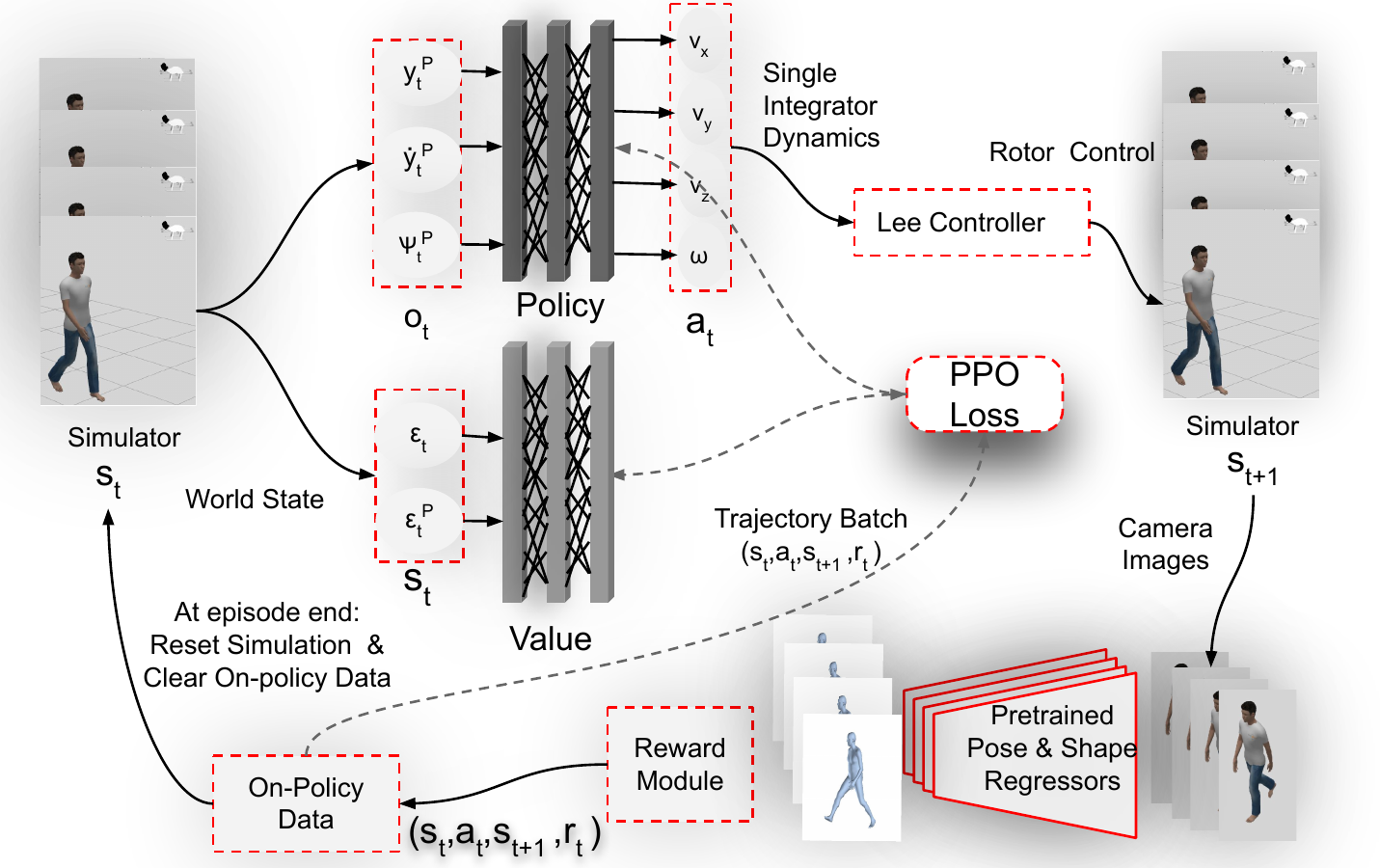}
 \caption{Single Agent Network: Variants of this network are trained with different rewards as described in sub-subsection \ref{subsec:proposed_method}--1.}
 \label{fig:singleagentnet}
\end{figure}

\paragraph{Network 2.2: Centering, collision avoidance and Multiview HMR Reward (Trained with Static Subject)}
In this variant we use a sum of three rewards $r_{\textrm{center}}$, $r_{\textrm{col}}$ and $r_{\textrm{MHMR}}$. The first two are same as (\ref{eqn:centrtingreward}) and (\ref{eqn:colreward}), respectively. $r_{\textrm{MHMR}}$ rewards the agent based on the output accuracy of the MoCap back end using images from multiple agents. For this, we use MultiviewHMR \cite{liang2019shape}. It is a state-of-the-art method for human pose and shape estimation using images from multiple viewpoints. At every timestep of training, we use it on the image acquired by the agent and its neighbor to compute an estimate of $\mathbf{\check{x}}_{j,t}^P \forall j; j=1\cdots14$ corresponding to all 14 joints. The reward is then given by
\begin{equation}
 r_{\textrm{MHMR}} = 1 - \tanh({c_4d_{\mathrm{mhmr}}}),
 \label{eqn:mviewhmr}
\end{equation}where $d_{\mathrm{mhmr}} = \frac{1}{14}\sum_{j=1}^{14}(w_j||\mathbf{\check{x}}_{j,t}^P  - \mathbf{\bar{x}}_{j,t}^P||_2)$ 
and the weights are as described in the previous section.

\paragraph{Network 2.3: Centering, continuous collision avoidance and Multiview HMR Reward (Trained with Moving Subject)}
In this variant we use a sum of three rewards $r_{\textrm{center}}$, $r_{\textrm{concol}}$ and $r_{\textrm{MHMR}}$. Here $r_{\textrm{center}}$ and $r_{\textrm{MHMR}}$ are same as (\ref{eqn:centrtingreward}) and (\ref{eqn:mviewhmr}). The continuous collision avoidance reward is given as follows.
\begin{equation}
r_{\textrm{concol}} = \begin{cases}
-v_{\textrm{pot}} ,& \text{if } \|\mathbf{x}_t-\bar{\mathbf{x}}_t\|_2\leq \mathbf{d}_{lthresh}\\
0.2 ,& \text{if } \mathbf{d}_{lthresh} \leq \|\mathbf{x}_t-\bar{\mathbf{x}}_t\|_2\leq \mathbf{d}_{hthresh}\\
-1 ,              & \text{otherwise}
\end{cases}
\label{eqn:concolreward}
\end{equation}where $\mathbf{d}_{lthresh} = 1.0$m and $\mathbf{d}_{hthresh} = 20$m. $v_{\textrm{pot}}$ is obtained using the potential field functions as described in our previous work \cite{rahul_CASE_2019} (equation 3). Furthermore, the value of $v_{\textrm{pot}}$ is clamped to $1$.

\paragraph{Network 2.4 + Potential Field: Centering and Multiview HMR Reward (Trained with Moving Subject)}
In this variant, we use a sum of two rewards, namely, $r_{\textrm{center}}$ (\ref{eqn:centrtingreward}) and $r_{\textrm{MHMR}}$ (\ref{eqn:mviewhmr}). The key difference in this case w.r.t. Network 2.3 is that here we use a potential field-based collision avoidance method \cite{rahul_CASE_2019} as a part of the environment during the training to keep the robots from colliding with each other at all times. It is not embedded in the reward structure and hence, the robots are not explicitly penalized for it. Testing of this network, during experiments, was also performed with potential field-based collision avoidance as a part of the environment.




%

\begin{figure}[!t]
 \includegraphics[width=0.9\columnwidth]{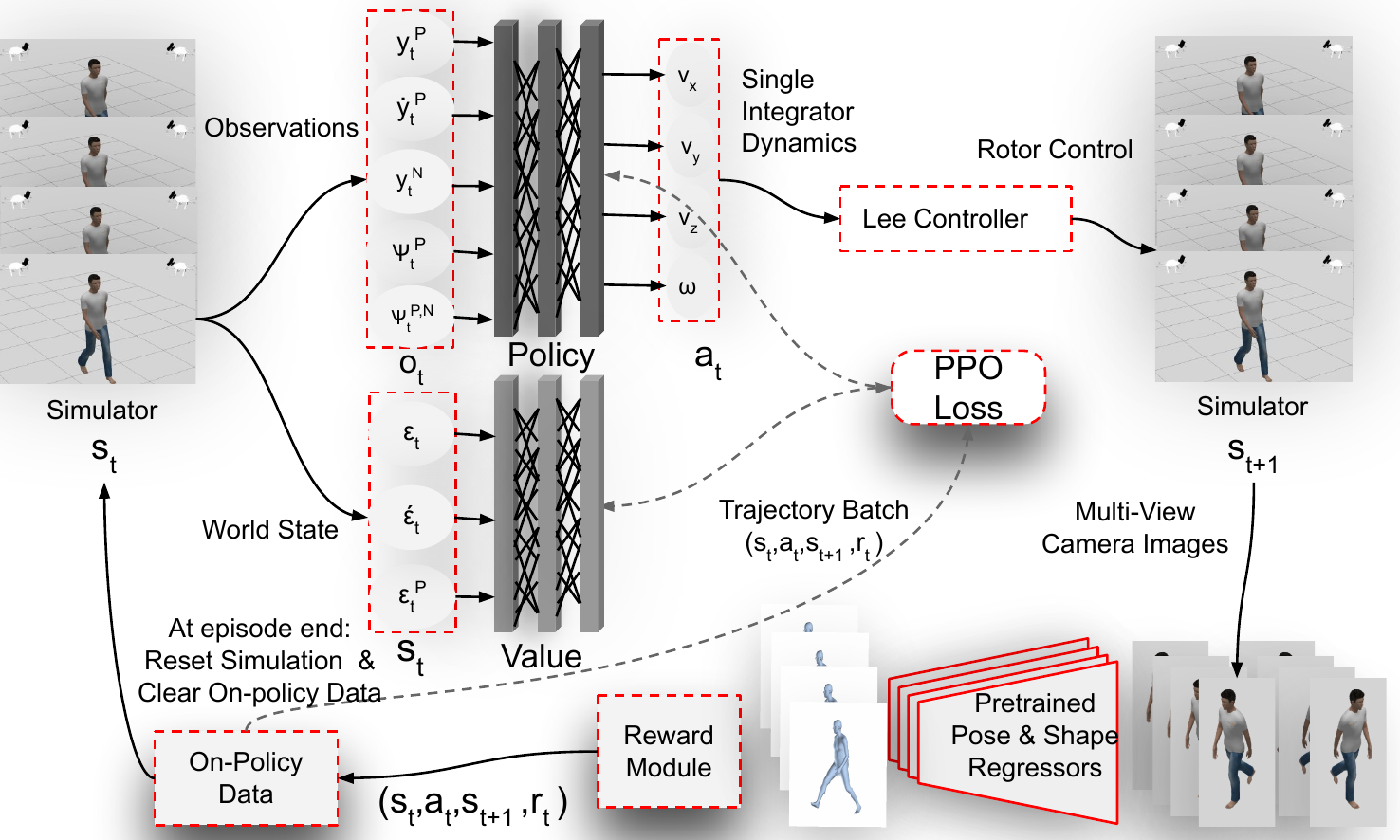}
 \caption{Multi Agent Network: Variants of this network are trained with different rewards as described in sub-subsection \ref{subsec:proposed_method}--2.}
 \label{fig:multiagentnet}
\end{figure}

\section{Experiments and Results}

\subsubsection{Training Setup in Simulation}


We train and our networks in simulation. We use Gazebo multi-body dynamics simulator with ROS and OpenAI-Gym to train the MAV agents. For the MAV agent we use AscTec Firefly model with an on-board RGB camera facing down at 45$^\circ$ pitch angle w.r.t.\ the MAV body frame. We run 5 parallel instances of Gazebo and the Alphapose network on multiple computers over a network to render the simulation. The policy network is trained on a dedicated PC which samples a batch of transition and reward tuples from the network of computers to update the networks. We use a simulated human in Gazebo as the MoCap subject and generate random trajectories using a custom plugin. Details of the network architectures, training process, libraries, instructions on how to run the code, etc., are provided in the attached supplementary material.

\subsection{Simulation results}

In this sub-section we evaluate our trained policies in Gazebo simulation environment. We create a test trajectory for the simulated human actor for $120$s on which it walks with varying speeds. The best policy of each network variant, as described in subsection~\ref{subsec:proposed_method}, is run 20 times while the actor walks the trajectory. Thus, results from a total of $2400$s of evaluation run of each network variant is obtained. 

For single agent experiments, in addition to the DRL-based methods, we run 4 other methods: i) `Network 1.4 + AirCap', ii) Orbiting Strategy, iii) Frontal-view Strategy and iv) MPC-based approach \cite{ActiveTallamraju19}. For multi-agent experiments we run 2 additional methods: i) `Network 2.3 + AirCap' and ii) MPC-based approach \cite{ActiveTallamraju19}. All these were also run 20 times for 120s each to allow comparison with our DRL-based policies. 
`Network 1.4 + AirCap' and `Network 2.3 + AirCap' imply running the networks with `true observations' instead of directly using simulator-generated ground-truth observations. To this end, we ran the complete AirCap pipeline \cite{ActiveTallamraju19} during the test by replacing only the MPC-based high-level controller with the DRL policy in it. It executes an NN-based person detector, a Kalman filter-based estimator for person's 3D position estimation (not orientation), cooperative self-localization of the MAVs using simulated GPS measurements with noise as well as communication packet loss. More details regarding this are provided in the supplementary material associated with this article.
`Orbiting Strategy' is essentially a `model-free' approach in which a robot orbits around the person at a fixed distance in order to increase the coverage. In `Frontal-View Strategy' a robot maintains a fixed distance to the person and attempts to always keep the frontal view of the person in the camera image.
Below we discuss the results for single and multi-agent network variants and other aforementioned methods.

\begin{figure}[t]
 \includegraphics[width=\columnwidth]{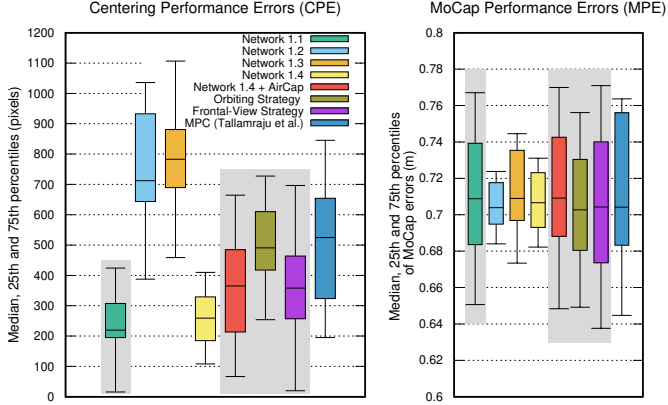}
 \caption{Simulation results of Single Agent Network variants.}
 \label{fig:sim_experiments_single_agent}
\end{figure}



\subsubsection{Single Agent Network Variants}

In order to compare the network variants, we use 2 metrics, i) centering performance error (CPE) and ii) MoCap performance error (MPE). CPE is computed as the pixel distance from the center of the bounding box around the person in the agent's camera image to the image center. MPE, for single agent networks is simply $d_{\mathrm{J}}$, as defined for the reward in (\ref{eqn:spinreward}). 
To compute this, the SPIN method \cite{kolotouros2019spin} is run on the images acquired by the agents during testing.

\begin{figure}[t]
 \includegraphics[width=\columnwidth]{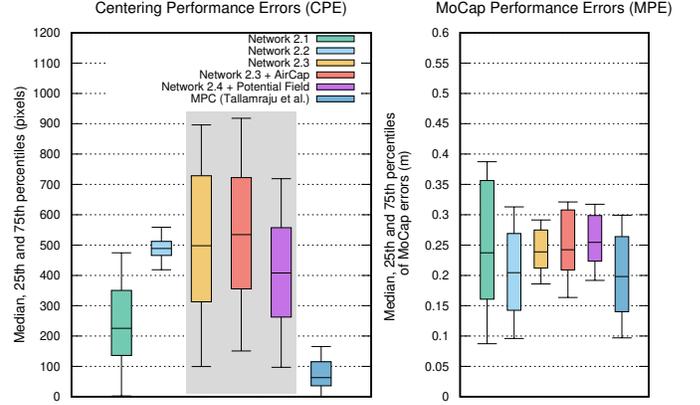}
 \caption{Simulation results of Multi-Agent Network variants.}
 \label{fig:sim_experiments_multi_agent}
\end{figure}

Note that the metric which quantifies the MoCap accuracy of any method in this paper is MPE (the right side box plots in Fig.~\ref{fig:sim_experiments_single_agent}  and \ref{fig:sim_experiments_multi_agent}). CPE is a metric that we plot only to make the policy performance intuitively explainable and understand `what' the learned RL policies are doing to achieve a good MPE.

Figure~\ref{fig:sim_experiments_single_agent} shows the error statistics of the aforementioned metrics. The grey background behind any box plot signifies that the method could not keep the person, even partially, in the MAV FOV, thereby completely losing him/her, for at least some duration of the experiment runs. In these cases, the box plot represents errors computed only for those timesteps when the person was at least partially in the FOV.


MPE plots in Fig.~5 for single robot experiments show that for all methods the medians of the MPEs are very similar to each other. This is the most significant result, especially because we can demonstrate that in terms of accuracy our DRL-based approach is on par with the state-of-the-art MPC-based approach [2] (or fixed-strategy methods), without the need for hand-crafting observation models and system dynamics (or pre-specified robot trajectories). Furthermore, the MPE for network 1.4 and 1.2 also has significantly less variance of MPE compared to all other methods. Due to these reasons, Network 1.4 and Network 1.2 are the two most successful approaches for the MoCap task.

From Fig.~5 plots, we also see that Network 1.4 keeps the person centered much more than Network 1.2, 1.3 or MPC. This is expected because Network 1.4 is rewarded for centering the person in the image in addition to SPIN-based MoCap rewards. Network 1.2 or 1.3, on the other hand, only has SPIN-based MoCap rewards. Nevertheless, the MPE of Network 1.4 is only slightly better than that of Network 1.3. This signifies that centering the person in the image does not have a great impact on the accuracy of the motion capture estimates.

Network 1.1, which often lost the person in its FOV, outperforms all other methods in its CPE performance for the duration it could `see' the person. This is expected as it is trained with only centering reward. Even though its MPE mean for the person-visible duration is similar to other networks, the variance of its MPE is higher than the other networks. Moreover, the fact that it could keep the person in FOV only $76\%$ of the time as compared to $100\%$ for other networks (1.2--1.4) makes it less desirable even for the MoCap task. 

The median MPE of `Network 1.4 + AirCap' is very similar to all other methods. However, it should be noted that there is one drawback in `Network 1.4 + AirCap'. As the `ground truth observations' are not used in this method and the simulated person can rapidly make sudden direction changes, the person is much more susceptible to go out of the FOV of the MAV's camera. Since the network never learned to `search' for the person who is out of the FOV, the method has to `wait' until the person walks back in the FOV. The cooperative estimation method of the AirCap pipeline helps in this regard as the person might still be in another robot's FOV. For a single robot case this is also not possible. Thus, `Network 1.4 + AirCap' loses the person for 35\% of the time.

The strategy-based methods struggle to keep the person, even partially, in the MAV camera's FOV. While the `Orbiting Strategy' was able to keep the person in the FOV for 73\% of the total time of all experiments combined, the `Frontal-View Strategy' managed to do that only 20\% of the total time. This is because when the person changes his direction or speed of motion, the robot could fly around to reposition itself in the front of the person, thus losing him during the transition. On the other hand, our successful DRL-based approaches, i.e., Network 1.2, 1.3 and 1.4, never lose the person from the camera FOV.
Based on this analysis, we can conclude that the strategy-based methods, while being `model-free', still have a major drawback of losing the person often, if not very carefully hand-crafted. Our DRL-based approaches `explore' the space of these strategies and finds the most suitable one in their policies.

\subsubsection{Multi-Agent Network Variants}

The MPE in the multi-agent case is also simply $d_{\mathrm{J}}$, as defined for the reward in (\ref{eqn:spinreward}), but instead of using SPIN as in the single agent case, here it is computed by running Multiview HMR \cite{liang2019shape} for pose and shape estimation on every simultaneous pair of images acquired by both the agents during the evaluation runs.
Network 2.1 and 2.2 were trained and tested on a static person. On the other hand, Network 2.3 and Network 2.4 + Potential Field were both trained and tested with a moving person (in the same way as for the single agent experiments). The remaining two methods in the multi-agent case were also tested with moving persons.



Figure~\ref{fig:sim_experiments_multi_agent} shows the error statistics of multi-agent simulation experiments. The best performing network in multi-agent case is Network 2.3. It is very similar to the MPC-based method in terms of the MPE median value (See Fig.~6 right side) and has much less MPE variance than MPC. This is a very significant result as MPC required observation models of the subject and our DRL-based approach in Network 2.3 did not. In the MPC approach, the viewpoint configurations for the MAVs emerge out of the joint target perception models. In contrast, in the DRL-based approach the MAVs directly learn the viewpoint configurations from experience. We also notice that the rewards based on a triangulation method assist, to some extent, in achieving acceptable MoCap performance (see results of Network 2.1). However, they remain inferior to the Network 2.3, which used the sophisticated approach taken in Multiview HMR \cite{liang2019shape} for reward computation.


Furthermore, we find that in terms of MPE, `Network 2.3 + AirCap' is close to both Network 2.3 and MPC. Similar to `Network 1.4 + AirCap', the `Network 2.3 + AirCap' also loses the person from the robots' FOV. However, it is present in at least one robot's FOV for approx.~97\% of the total experiment duration. The increased visibility in the multi-robot case is due to the cooperative estimator module of AirCap pipeline. This assessment signifies the usability of our method in real robots with real observations.

Next, we find that the policy learned by `Network 2.4 + Potential field' was able to achieve MPE median value comparable to Network 2.3 but at the cost of slightly higher MPE variance and loss of person from at least one robot's FOV for several periods (13\% of total duration). This experiment further signifies the key benefit of our DRL-based approach in Network 2.3. It overcomes the need for knowing models, strategies as well as any ad-hoc collision avoidance techniques. In Network 2.3 the learned policy not only achieves good MoCap performance, but it also naturally learns to avoid collisions with the teammates. In the video associated to this paper (also available here -- \url{https://youtu.be/07KwNjc7Sy0}) we show how well Network 2.3 performs.
The networks for the moving person, however, did not ensure very good centering of the person in the image (see the left side of Fig.~6) as compared to the MPC-based approach. Despite this, their MPE performances are only slightly poorer than MPC (MPE median difference is approx.~0.05m only). This further signifies that centering the person on the image has a very low effect on MoCap performance.


\begin{figure}[t]
\centering
 \includegraphics[width=0.9\columnwidth]{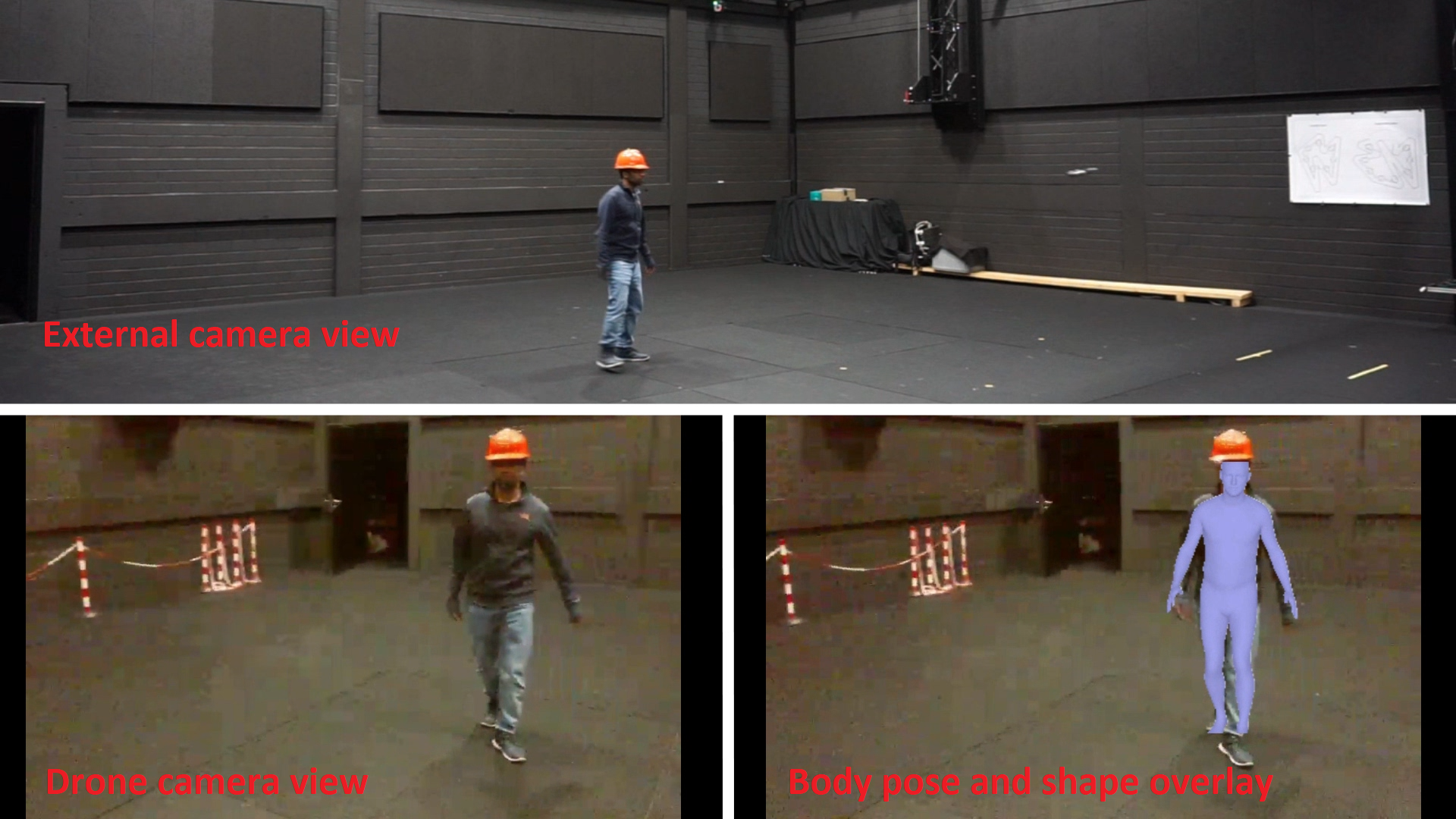}
 \caption{A snapshot of the real robot experiment.}
 \label{fig:realrobotexpsfootage}
\end{figure}

Finally, for the multi-agent case, we find that the medians of the MPEs for all multi-agent networks were substantially lowered compared to the MPEs obtained by single-drone experiments (from $\sim$ 0.7m to 0.22m). This highlights the benefit of using multiple drones and hence multiple views to improve MoCap performance.

\subsection{Real Robot results}



In order to validate our approach in a real robot scenario, we used a DJI Ryze Tello drone. It consists of a forward looking camera capturing images at $30$ hz. The drone is controllable using an SDK with ROS interface. Tello has the functionality of vision-based localization, which is highly inaccurate. Hence, we performed experiments within a Vicon hall with markers on top of the drone to estimate its position and velocity. The tracked subject wore a helmet with Vicon markers. Vicon-based position estimate of the person was used to compute the observations for the neural network.

\begin{figure}[!t]
 \includegraphics[width=\columnwidth]{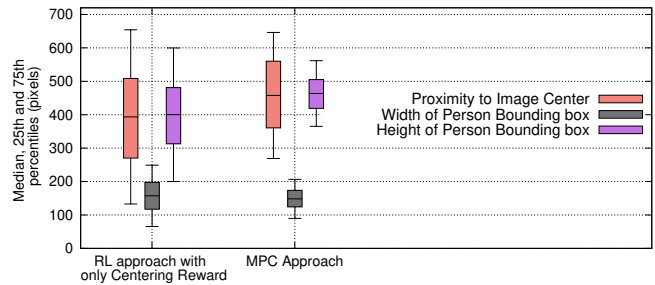}
 \caption{Real Robot Experiments: Comparison of single agent network variant 1.1 and MPC-based \cite{ActiveTallamraju19} approach.}
 \label{fig:realrobotexps}
\end{figure}

We performed experiments with $1$ Tello drone and compared our DRL-based approach using Network 1.1 with state-of-the-art MPC-based approach \cite{ActiveTallamraju19}. These were performed for approximately $400$s and $700$s, respectively. Figure~\ref{fig:realrobotexpsfootage} shows an external camera footage of the experiment and the on-board drone view with pose and shape overlay using SPIN. As the ground truth pose and shape of the human subject in real experiment is not available, we only compare the following criteria. We compare i) the length and breadth of the bounding box around the person in the drone images, and ii) proximity of the person to the center of those images, calculated as pixel distance from the image center to the center of the bounding box around the person. The bounding boxes are computed by running Alphapose \cite{cao2017realtime} method on the images recorded by the drone. Figure~\ref{fig:realrobotexps} presents the statistics of these evaluation criteria. We notice that the performance of both approaches is similar in terms of the person's proximity to the image center, with our DRL-based approach performing slightly better. However, we observe that the MPC-based approach is consistently able to keep a larger size (projected height) of the person in the images. This is due the fact that the MPC's objectives enforce it to keep a certain threshold distance to the person. As the DRL-based approach has no such incentive, it varies its distance to the person more, therefore causing a greater variance in the projected height of the person. On the other hand, this enables our DRL-based approach to change its relative orientation with respect to the person such that she/he is is observed from several possible sides. This is evident by the greater variance in the projected width of the person on the images. This property of our DRL-based approach will benefit pose and shape estimation methods, as demonstrated in the simulation experiments.

\section{Conclusions and Future Work}

In this letter, we presented the first deep reinforcement learning-based approach to human motion capture using aerial robots. 
Our solution does not depend on hand-crafted system or observation models. Formation control policies are directly learned through experience, which is obtained in synthetic training environments. 
Through extensive experiments and comparisons we find that DRL-based agents learn extremely good policies, on par with carefully designed model-based (MPC) or model-free, fixed strategy-based methods. These policies even generalize to real robot scenarios. We also find that multiple agents learn even better policies and outperform single agents in performing MoCap.
\
The learning objective (MoCap accuracy) is far simpler to construct than deriving system or observation models \cite{ActiveTallamraju19}. Moreover, strategy based methods, as shown in our experiments, can have various drawbacks, such as losing the person from the field of view. To overcome that, each drawback must be identified and addressed within the fixed strategy. A DRL-based approach overcomes the need for fixing a strategy by `exploring' the space of such strategies. Thus, a major conclusion of our work is that DRL-based approaches are likely the ideal way forward for aerial MoCap systems. Eventually, an end-to-end approach of learning actions directly from images is needed to overcome the need for an additional person-detection method, that has been used so far (e.g., SSD multibox in AirCap \cite{DeepPrice18}). To this end, we are improving our training by using SMPL body models in richer, photorealistic simulated environments.

Our approach would also be applicable in a real robot setting with `real observations' while achieving accuracy similar to an MPC-based approach \cite{ActiveTallamraju19}. Nevertheless, this is valid only for those durations when the person is not lost from the FOV of all cameras. In order for the policy to `search' for the person, network training should be done with the AirCap pipeline's `real observations'. This would involve running several DNN-based detectors and keeping track of delayed measurements. Furthermore, our approach is limited in terms of scaling up to more agents. While addressing this will require more sophisticated network architecture, it should be noted that 2 to 3 aerial robots may be enough to achieve a good MoCap accuracy \cite{MarkerlessNitin19}.


\section*{Acknowledgments}
The authors would like to thank Prof.\ Dr.\ Heinrich B\"ulthoff for his constant support and providing us the access to the Vicon tracking hall in MPI for Biological Cybernetics. The authors also thank Igor Martinovi\'c and the anonymous reviewers for extremely helpful suggestions.

\bibliographystyle{IEEEtran}  
\bibliography{bibliography}

\end{document}